\documentclass[runningheads]{llncs}
\usepackage{makeidx}
\usepackage{graphicx}
\usepackage{amsmath,amssymb} % define this before the line numbering.
\usepackage{color}
\usepackage{multirow}
\usepackage{booktabs}
\usepackage{hyperref}
\usepackage{upgreek}
\usepackage{bm}
\usepackage{setspace}
\definecolor{ocre}{RGB}{0,0,102}
\usepackage{caption}
\usepackage[font={color=ocre,small}]{caption}
\usepackage{pifont}% http://ctan.org/pkg/pifont
\newcommand{\STAB}[1]{\begin{tabular}{@{}c@{}}#1\end{tabular}}

\def\eg{\emph{e.g. }}

\def\ie{\emph{i.e. }}
\def\etal{\emph{et al.}}

\begin{document}
\title{Vehicle Re-Identification in Context} 
% Replace with your title

\titlerunning{Vehicle Re-Identification in Context}
% Replace with a meaningful short version of your title
%
\author{Ayta\c{c} Kanac\i\inst{1} \and
Xiatian Zhu\inst{2} \and
Shaogang Gong\inst{1}}
%
%Please write out author names in full in the paper, i.e. full given and family names. 
%If any authors have names that can be parsed into FirstName LastName in multiple ways, please include the correct parsing, in a comment to the volume editors:
%\index{Lastnames, Firstnames}
%(Do not uncomment it, because you may introduce extra index items if you do that, we will use scripts for introducing index entries...)
\authorrunning{Kanac\i, A. and
Zhu, X. and
Gong, S.}
% Replace with shorter version of the author list. If there are more authors than fits a line, please use A. Author et al.
%

\institute{Queen Mary University of London, London E1 4NS, UK \and 
Vision Semantics Limited, London E1 4NS, UK\\
\email{\{a.kanaci,s.gong\}@qmul.ac.uk}\\
\email{eddy@visionsemantics.com}}
\maketitle              % typeset the header of the contribution

\begin{abstract}
Existing vehicle re-identification (re-id) evaluation benchmarks
consider strongly artificial test scenarios by assuming the availability of
high quality images and fine-grained appearance at an almost constant
image scale, reminiscent to images required for Automatic Number Plate
Recognition, e.g. VeRi-776. 
Such assumptions are often invalid in realistic vehicle re-id
scenarios where arbitrarily changing image resolutions (scales) are
the norm. This makes the existing vehicle re-id 
benchmarks limited for testing the true performance of a re-id method. 
In this work, we introduce a more realistic and challenging 
vehicle re-id benchmark, called Vehicle Re-Identification in Context (VRIC).
In contrast to existing vehicle re-id datasets,
VRIC is uniquely characterised by vehicle images
subject to more realistic and unconstrained variations in
resolution (scale), motion blur, illumination,
occlusion, and viewpoint. %, and background clutter.
It contains 60,430 images of 5,622 vehicle identities captured by 
60 different cameras at heterogeneous road traffic scenes
in both day-time and night-time.
Given the nature of this new benchmark, we further investigate a
multi-scale matching approach to vehicle re-id by learning more
discriminative feature representations from multi-resolution images.
Extensive evaluations show that the proposed multi-scale method
outperforms the state-of-the-art vehicle re-id methods on three
benchmark datasets: VehicleID, VeRi-776, and VRIC\footnote{Avaliable at \url{http://qmul-vric.github.io}}.
\end{abstract}

%-------------------------------------------------------------------------
\section{Introduction}
\label{sec:intro}
Vehicle re-identification (re-id) aims at searching vehicle 
instances across non-overlapping camera views
by image matching \cite{Liu2016cvpr_drdl}. 
Influenced by the recent extensive studies on person re-id
\cite{gong2014person,xiao2016learning,wang2014person,li2014deepreid,zhu2018fast,sun2017svdnet,li2018Unsupervised,wang2018person,zhong2017camera}, 
vehicle re-id has started to gain increasing attention in the past two
years, which promises the potential for more flexible means for
vehicle recognition and search than Automatic Number Plate Recognition
(ANPR).	
However, vehicle re-id by visual appearance is a challenging task
due to the very similar appearance of different vehicle instances
of the same model type and colour, and a significant visual appearance
variation of the same vehicle instance in different camera views. 

\begin{figure}[ht]
    \centering
    \begin{tabular}{ccc}
        {\small VehicleID} &{\small VeRi-776 } & {\small VRIC } \\
        \begin{minipage}{.3\textwidth}
              \includegraphics[width=\linewidth]{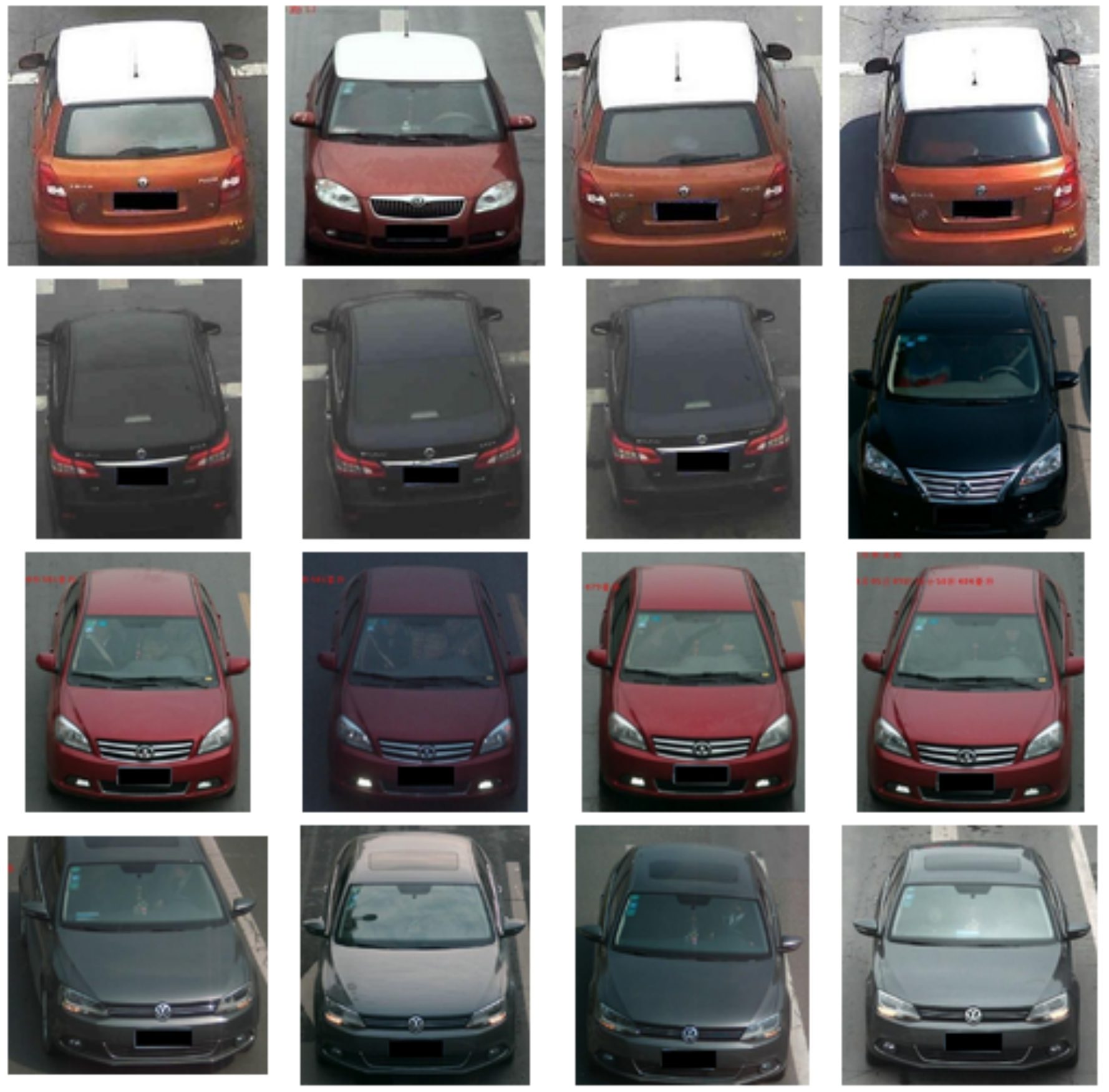}
        \end{minipage}
        &
        \begin{minipage}{.3\textwidth}
              \includegraphics[width=\linewidth]{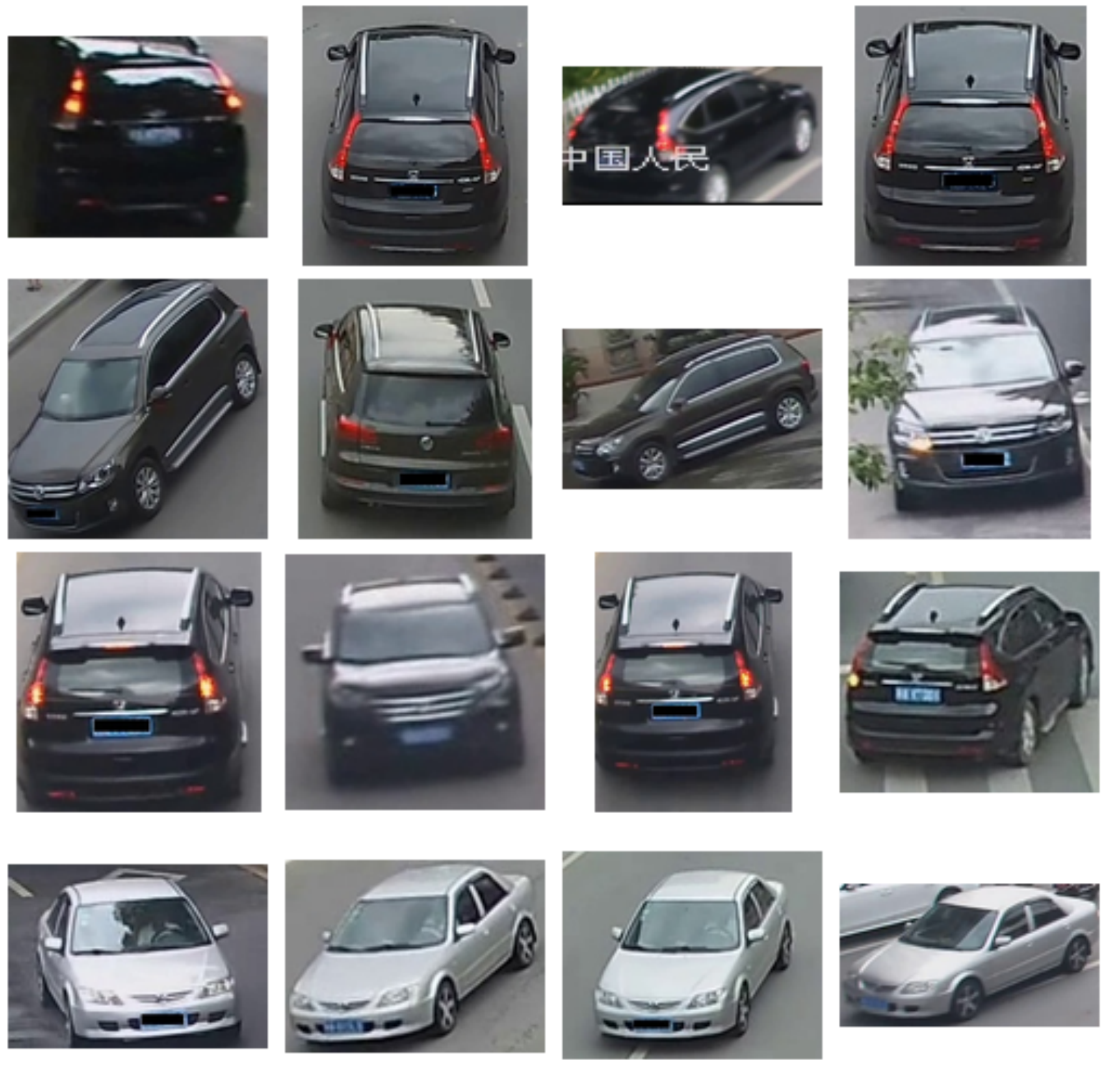}
        \end{minipage}
        &
        \begin{minipage}{.3\textwidth}
              \includegraphics[width=\linewidth]{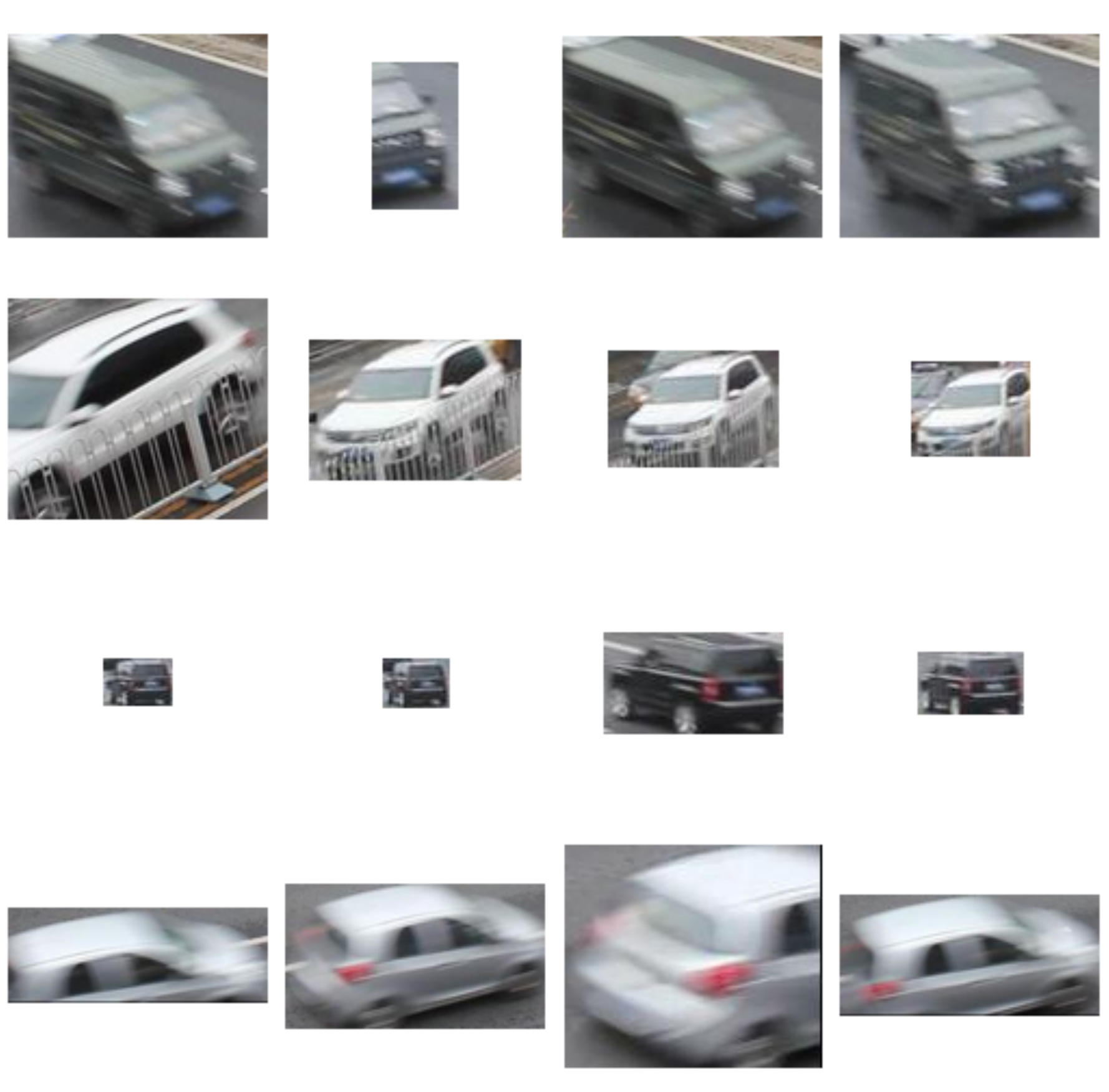}
        \end{minipage}
        \\

    \end{tabular}
%	\vspace{-0.2cm}
    \caption{
    	Example images of VehicleID, VeRi-776 and VRIC.
    	Images in each row depict the same vehicle instance. 
    	VRIC images exhibit significantly more unconstrained variations 
    	in resolution, motion blur, occlusion/truncation and illumination 
    	within each vehicle bounding-box images.
%    	Example images from the datasets highlighting the multi resolution and motion artifacts of the newly introduced Detrac4reID. Each row of images are from certain identity.
    }
    \label{fig:dataset_img}
%    \vspace{-0.7cm}
\end{figure}

Current vehicle re-id studies are mainly driven by
two benchmark datasets, VehicleID \cite{Liu2016cvpr_drdl}
and VeRi-776 \cite{Liu2016eccv_veri}.
While having achieved significant performance improvement 
(e.g. from 61.44\% by \cite{Liu2016eccv_veri} to 92.35\% Rank-1 by \cite{Wang_2017_ICCV} on VeRi-776),
the scalability of existing re-id algorithms to real-world vehicle
re-id applications remains unclear. 
This is because existing benchmarks represent somewhat rather artificial
tests using high-quality images of
high resolution, no motion blur, limited weather conditions and
occlusion %against non-cluttered background 
(Table \ref{tab:datasets} and Fig \ref{fig:dataset_img}). This is more
reminiscent to imaging conditions for ANPR than what is typical for vehicle
re-id in wide-view traffic scenes ``in-the-wild''.
%In particular, current models trained by such artificially
%high-quality vehicle images may not be robust for generalisation to
%more realistic vehicle re-id conditions. 
%
%To address this problem, it is crucial to introduce 
%a dataset that matches the realistic deployment settings.

In this work, we introduce a new benchmark dataset called {\bf Vehicle
  Re-Identification in Context} (VRIC) for more realistic and challenging
vehicle re-identification.
VRIC consists of 60,430 images of 5,656 vehicle IDs collected from 60 different cameras
in traffic scenes. 
VRIC differs significantly from existing datasets in 
that {\em unconstrained} vehicle appearances were captured
with variations in imaging resolution, motion blur,
weather condition, and occlusion.
% and background clutter.
%
%These are necessary and essential characteristics of a realistic test that inevitably impose more significant challenges to competing algorithms for evaluating their performance robustness in reality.
%
This VRIC dataset aims to provide a more realistic 
vehicle re-id evaluation benchmark.
%
%All vehicles of this dataset are in operation,
%so we call it {\em }
%

We make two contributions:
(1) We create and introduce a more realistic vehicle re-id benchmark
VRIC that contains vehicle images of {\em unconstrained} visual
appearances with variations in resolution, motion blur,
weather setting, and occlusion. % and background clutter.
This dataset is created from the UA-DETRAC benchmark \cite{detrac_wen}
originally designed for object detection
and multi-object tracking in traffic scenes, therefore reflecting
appropriately and providing the necessary vehicle re-id environmental
context and viewing conditions.
This new benchmark will be publicly released.
(2) We further investigate a Multi-Scale (resolution) Vehicle Feature
(MSVF) learning model to address the inherent and significant
multi-scale resolution in vehicle visual appearances from typical
wide-view traffic scenes, currently an unaddressed problem in vehicle
re-id due to the lack of a suitable benchmark dataset.
Extensive comparative evaluations demonstrate the effectiveness of the
proposed MSVF method in comparison to the state-of-the-art vehicle
re-id techniques on the two existing benchmarks (VehicleID
\cite{Liu2016cvpr_drdl} and VeRi-776 \cite{Liu2016eccv_veri}) and the
newly introduced VRIC benchmark.

\begin{table}  
%	\footnotesize
	\centering
	\setlength{\tabcolsep}{0.05cm}
	\caption{
		Characteristics of vehicle re-id
                datasets. 
%        The ``Mean Resolution'' value is in pixels.
		% Resolution is measured by \texttt{max}(width, height).
		% {\bf {\color{red}TODO: fill missing numbers.}}
		% cloudy, night, sunny, and rainy
	}
%	\vspace{-0.2cm}
    \resizebox{\columnwidth}{!}{%
	\label{tab:datasets}
	\begin{tabular}{c||c|c|c|c|c|c|c}
		\hline
		Dataset   
        & Images & IDs & Cameras &  \begin{tabular}[c]{@{}c@{}} Resolutions\\Width$\times$Height (Mean)\end{tabular} & Motion Blur & Illumination & Occlusion 
        %& Background 
        \\
		\hline \hline                                                                                                                      
		VehicleID \cite{Liu2016cvpr_drdl}   
        & 113,123 & 15,524 & - & 345.4$\times$376.1        & No       & Limited & No 
        %& Uniform 
        \\ 
		\hline
		VeRi-776 \cite{Liu2016eccv_veri}       
        & 51,034 &  776 & 20 &  376.1$\times$345.4     & No   & Limited  & No 
        %& Uniform  
        \\ 
		\hline
		VD1 \cite{yan2017exploiting} 
        & 846,358 & 141,756& - & 424.8$\times$411.0      & No       & Limited & No 
        %& Uniform
		\\ \hline
		VD2 \cite{yan2017exploiting} 
        &  690,518 &  79,763 &- & 401.3$\times$376.4       & No       & Limited & No %& Uniform
		\\ \hline \hline
		{\bf VRIC} ({Ours})                       
        & 60,430 &  5,622  & 120 & 65.9$\times$103.0 & Unconstrained  & Unconstrained  & Unconstrained %& Cluttered 
        \\   
		\hline
	\end{tabular}
    }
%    \vspace{-0.5cm}
\end{table}

\section{Related Work}

\subsubsection{  Vehicle Re-Identification. }
Whist vehicle re-id is less studied than person re-id
\cite{gong2014person,li2014deepreid,ahmed2015improved,xiao2016learning,li2017person,sun2017svdnet,li2018harmonious,zhong2017camera,chen2018person,chen2018deep,wang2018reid},
there are a handful of existing methods.
Notably, Feris \etal \cite{feris2012large} proposed an attribute-based re-id method. 
The vehicles are firstly classified by different attributes like car model types and colours. The re-id matching is then conducted in the attribute space.
Dominik \etal \cite{zapletal2016vehicle}
used 3D bounding boxes for rectifying car images and then concatenate colour histogram features of vehicle image pairs. 
A binary linear SVM model is then trained to verify whether a pair of images have the same identity.
Both methods rely heavily on weak hand-crafted visual features in a
complex multi-step based approach, suffering from weak discriminative model generalisation.

More recently, deep learning techniques have been exploited to vehicle re-id.
Liu \etal \cite{Liu2016eccv_veri} explored a deep neural network to estimate the visual similarities between vehicle images.
%Vehicle appearances, spatio-temporal information and license plates are independently and incrementally used to improve the similarity function for vehicle matching. 
%they use a deep neural network to estimate the visual similarities between vehicle images
%for enjoying the strong representation learning capacity.
%
Liu \etal \cite{Liu2016cvpr_drdl} designed a Coupled Clusters
Loss (CCL) to boost a multi-branch CNN model for vehicle re-id. 
Kanaci \cite{kanaci2017vehicle} explored the
appearance difference at the coarse-grained vehicle model level. 
All these methods utilise the global appearance features of vehicle
images and ignore local discriminative regions.
To explore local information and motivated by the idea of landmark alignment \cite{zhang2014facial} 
in both face recognition \cite{taigman2014deepface} and human body pose estimation
\cite{newell2016stacked}, Wang \etal \cite{Wang_2017_ICCV} considered
20 vehicle keypoints for learning and aligning local regions of a vehicle for re-id.
Clearly, this approach comes with extra cost of exhaustively labelling these keypoints
in a large number of vehicle images, and the implicit assumption of
having sufficient image resolution/details for computing these keypoints.

Additionally, 
space-time contextual knowledge has also been exploited for vehicle re-id subject to structured scenes \cite{Liu2016eccv_veri,Shen_2017_ICCV}.
Liu \etal \cite{Liu2016eccv_veri} proposed
a spatio-temporal affinity approach for quantifying every pair of images.
%This method is inclined to image pairs that are close to each other 
%in both spatial and temporal domains therefore only a simplified solution.
%
Shen \etal \cite{Shen_2017_ICCV} further 
incorporated spatio-temporal path information of vehicles.
Whilst this method improves the re-id performance on the VeRi-776 dataset,
it may not generalise to complex scene structures when the number of 
visual spatio-temporal path proposals is very large with only weak contextual 
knowledge available to facilitate model decision.

In contrast to all existing methods as above,
we address a different problem of learning multi-scale feature
representation for vehicle re-id.

%In practice, this deep model however is not training-friendly due to 
%the need for complex parameter tuning in a multi-stage training process.

%Re-identification can be described as an extension of the classification problem
%where testing is done with unseen labels(identities) that are disjoint from the
%training set, with the caveat the all images belong to a certain type of
%``class'' \eg~person, vehicle. 

%\noindent \textbf{Fine grained model classification}
%With vehicle re-id, the hierarchical nature of
%vehicle categories also makes an interesting classification problem such as vehicle type, make, model
%and production year that is closely related. Yang~\etal~\cite{yang2015compcars}
%introduced large scale ``CompCars`` dataset with this purpose and shows CNN's
%perform decently with no bells and whistles in this context. 

%\noindent \textbf{Person Re-identification}
%Compared to person re-identification by either faces 
%\cite{huang2007labeled,guo2016ms,zhang2015beyond,nech2017level,liu2017sphereface,tran2017disentangled} or 
%whole bodies \cite{gong2014person,farenzena2010person,wang2014person,zheng2015partial,chen2017person,wang2016towards,wang2016human,li2017person,wang2016person,ma2017person,wang2016highly,peng2017joint}, 
%these methods were simple and brittle.

\subsubsection{Vehicle Re-Identification Benchmarks. }
There are in total four vehicle re-id benchmarks reported in the literature.
Liu~\etal~\cite{Liu2016cvpr_drdl} introduced the ``VehicleID'' benchmark 
with a total of 221,763 images from 26,267 IDs.  
In parallel, 
Liu~\etal~\cite{Liu2016icme-veri} created ``VeRi-776``, 
a smaller scale re-id dataset (51,035 images of 776 IDs) but with space-time annotations among 20 cameras in a road network.
Recently, Yan \etal \cite{yan2017exploiting} presented two larger datasets
(846,358 images of 141,756 IDs in ``VD1'', 690,518 images of 79,763 IDs
in ``VD2'')
with similar visual characteristics as VehicleID.

\begin{figure} %[ht]
    % \begin{minipage}{\textwidth}
    %     \centering
    %       \includegraphics[width=\linewidth]{imgs/detrac_train.pdf}
    % \end{minipage} \\
    % \begin{minipage}{.33\textwidth}
    %     \centering
    %       \includegraphics[width=\linewidth]{imgs/detrac_pg.pdf}
    % \end{minipage}
    % \begin{minipage}{.65\textwidth}
    %     \centering
    %     \includegraphics[width=\linewidth]{imgs/fig_weather.jpg} 
    % \end{minipage} \\
    % \begin{minipage}{\textwidth}
    %     \centering
    %       \includegraphics[width=\linewidth]{imgs/probes.pdf}
    %       \includegraphics[width=\linewidth]{imgs/gallerys.pdf}
    % \end{minipage} 
        \centering
        \includegraphics[width=\linewidth]{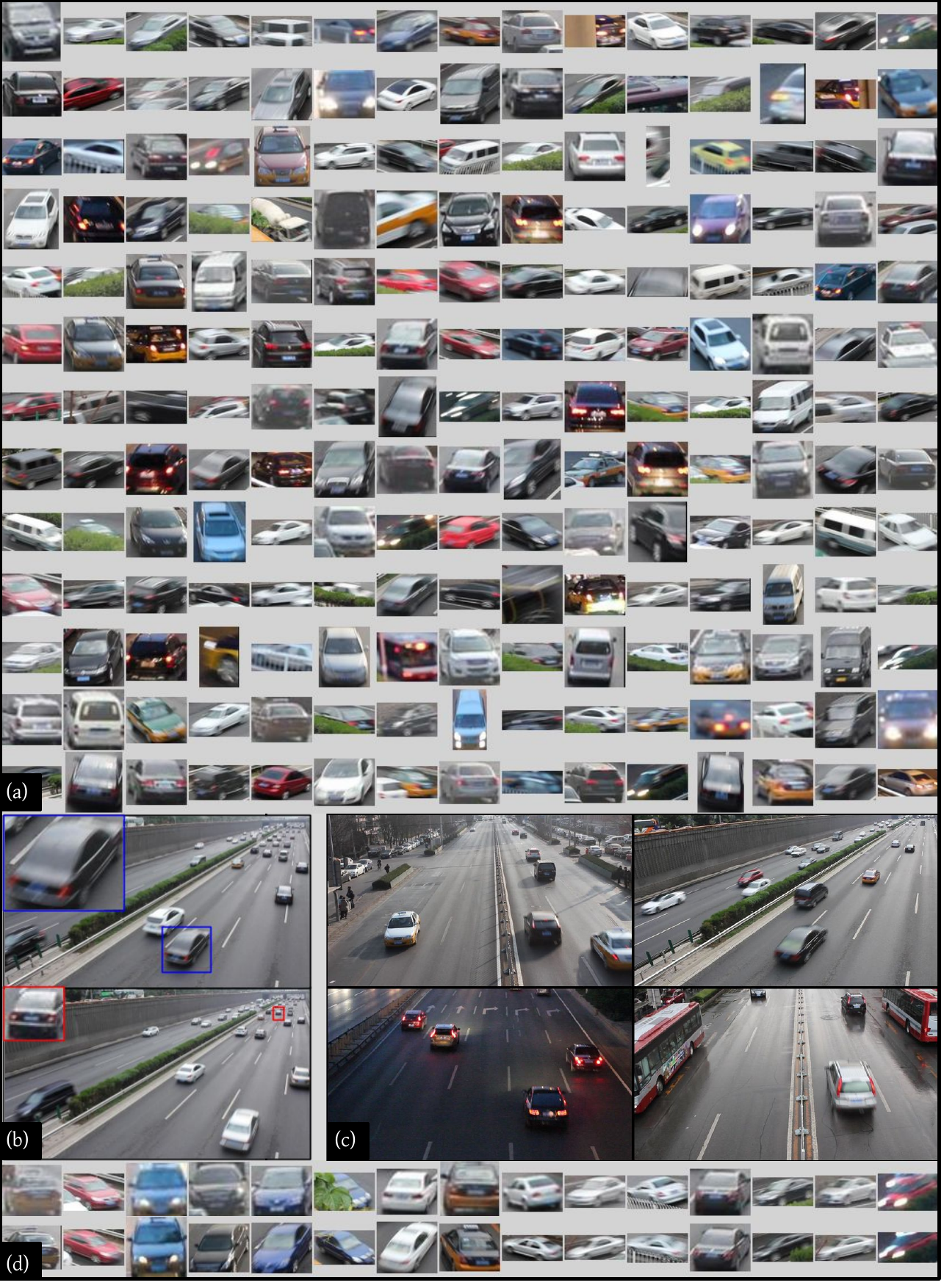}
%        \vspace{-0.2cm}
        \caption{Example vehicle bounding-box and whole scene images of the
    	VRIC benchmark.
    	{\bf (a)} Samples of vehicle bounding-box images.
    	{\bf (b)} The {\em near} and {\em far} views in a wide-view traffic scene.
    	{\bf (c)} UA-DETRAC video scenes with different illumination
        due to changing weather conditions (sunny, cloudy and rainy) and time (day and night).
    	{\bf (d)} Vehicle matching pairs (each column) from some
        example test vehicle instances. % in the test set.
    	%
%        (a) Example bounding box vehicle images from VRIC.
%        (b) Two examples frames from UA-DETRAC dataset and the probe
%        (red) and gallery (blue) images that extracted from these frames for
%        VRIC.
%        (c) Original frames from UA-DETRAC showing different illumination (weather) 
%        conditions. (Clockwise order: Sunny, Cloudy, Night, Rainy)
%        (d) Vertically matched probe and gallery image pairs.
}
   
    \label{fig:VRIC}
\end{figure}

Whilst these existing benchmarks have contributed significantly to the 
development of vehicle re-id methods, 
they only represent {\em constrained} test scenarios due to
the rather artificial assumption of having high quality images of
constant resolution (Table \ref{tab:datasets}).
This makes them limited for testing the true robustness of 
re-id matching algorithms in typically {\em unconstrained} wide-view traffic
scene imaging conditions.
The VRIC benchmark introduced in this work 
addresses this limitation by providing a vehicle re-id dataset
conditions giving rise to changes in resolution, 
motion blur, weather, illumination, and occlusion
(Fig \ref{fig:VRIC}). 

\section{The Vehicle Re-Identification in Context Benchmark}

\subsection{Dataset Construction}

We want to establish a realistic vehicle re-id evaluation benchmark with
natural visual appearance characteristics and matching challenges (Sec \ref{sec:intro}).
To this end, it is necessary to collect a large number of vehicle
images/videos from wide-view traffic scenes.
In the following, we describe the process of constructing the Vehicle
Re-Identification in Context (VRIC) benchmark.

\subsubsection{Source Video Data } 
Given highly restricted access permission of typical surveillance video data, 
we propose to reuse existing vehicle related datasets publicly available in the research community.

\begin{figure} %[t]{.4\linewidth}
	\centering
	\includegraphics[width=0.6\linewidth]{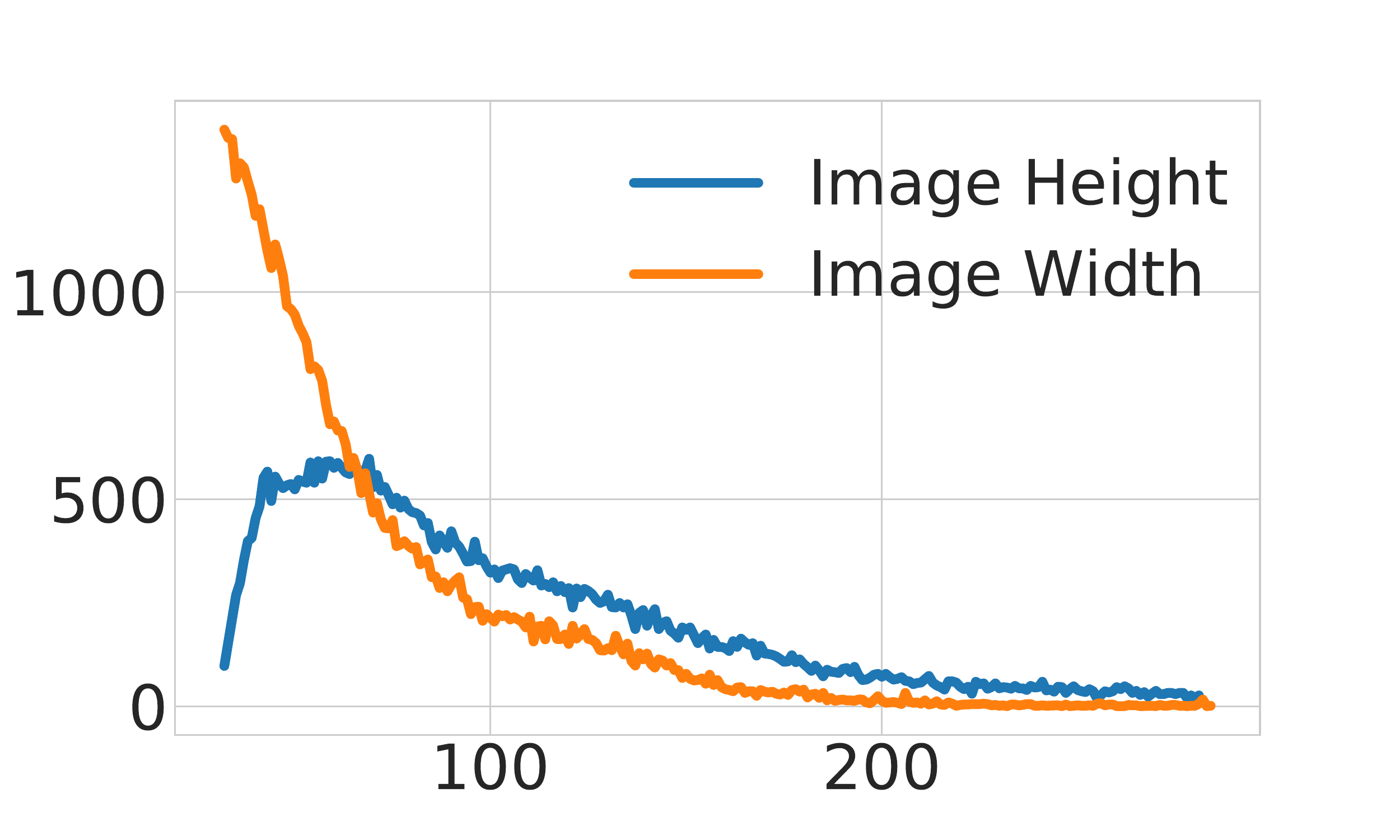}
%	\vspace{-0.2cm}
	\captionof{figure}{
		Vehicle instance scale distributions in VRIC.
		% {\bf \color{red}TODO: cut off the tail with 0. Enlarge font size of all text and numbers close to caption size.}
	}
	\label{fig:image_dim}
\end{figure}

In particular, we selected the UA-DETRAC
object detection and tracking benchmark \cite{detrac_wen} as the source data
of our VRIC benchmark, based on following considerations:
\begin{enumerate}
\item All videos were captured from the real-world traffic scenes
(e.g. roads), reflecting realistic context for vehicle re-id.

\item It covers 24 different surveillance locations with diverse environmental conditions therefore offering a rich spectrum of test scenarios
without bias towards particular viewing conditions.

\item It contains rich object and attribute annotations that can facilitate
vehicle re-id labelling.
\end{enumerate}
%(1) 
%
%(2) 
%
%(3) 
%
The UA-DETRAC videos were recorded at 25 frames per second (fps) with 
a frame resolution of 960$\times$540 pixels (Fig \ref{fig:image_dim}).
Samples of the whole scene images are shown in Fig \ref{fig:VRIC}(b,c).

%is a multi-object detection 
%and multi-object tracking and detection benchmark. 
%The training set consist of 60 real-world video recordings 
%of traffic with stationary cameras with resolution 960x540 at 25 fps. 
%It provides labelled bounding boxes and for 5952 vehicles for detection and
%tracking.

\subsubsection{Vehicle Image Filtering and Annotation. }
To construct a vehicle re-id dataset, 
we used 60 UA-DETRAC training videos with object bounding box annotations.
For vehicle identity (ID) annotation,
we started with assigning a unique label to each vehicle trajectory per UA-DETRAC video
and then manually verified the ID duplication cases.
Since all these raw videos were collected from different scenes 
and time durations, we found little duplicated trajectories
in terms of identity.
To ensure sufficient vehicle appearance variation,
we throw away short trajectories with less than 20 frames and 
bounding boxes smaller than 24$\times$24. 
By doing so, we obtained 5,622 vehicle IDs across all 60 videos.

In terms of vehicle instance resolution, 
the average image resolution of all 60,430 vehicle bounding-boxes is
69.8$\times$107.5 pixels in width$\times$height, with
a variance of 32 to 280 pixels due to the unconstrained distances
between vehicles and cameras. This presents inherently a multi-scale re-id matching challenge.

%In the spirit of other tracking dataset conversions such
%as~\cite{gou2017dukemtmc4reid} in
%person re-identification, we have transformed UA-DETRAC~\cite{detrac_wen}, 
%multi vehicle tracking benchmark, into a re-identification  dataset called 
%\textit{Detrac4reID}. 

%\begin{table}[ht]
%\begin{minipage}[b]{0.58\linewidth}
%	\centering
%        \caption{Data statistics and partition in VRIC.}
%        \vspace{0.2cm}
%        \setlength{\tabcolsep}{0.3cm}
%		\resizebox{\columnwidth}{!}{%
%        \begin{tabular}{l||c||c|c|c}
%            \hline 
%            \multirow{2}{*}{Partition} 
%            & \multirow{2}{*}{All}
%            & \multirow{2}{*}{Training Set} & \multicolumn{2}{c}{Test Set} \\
%            \cline{4-5}
%            & & & Probe & Gallery  \\
%            \hline \hline
%            IDs
%            & 5,622
%            & 2,813 & 2,813 & 2,813 \\
%            \hline
%            Images 
%            & 60,430 
%            & 54,808 & 2,813 & 2,813 \\
%            \hline
%        \end{tabular}
%		}
%        \label{table:data_split}
%\end{minipage}\hfill
%\begin{minipage}[t]{.4\linewidth}
%	\centering
%	\includegraphics[width=\linewidth]{hw_dist.pdf}
%	\captionof{figure}{
%	Scale distributions in VRIC.
%	% {\bf \color{red}TODO: cut off the tail with 0. Enlarge font size of all text and numbers close to caption size.}
%	}
%	\label{fig:image_dim}
%\end{minipage}
%\vspae{-.5cm}
%\end{table}

\subsection{Evaluation Protocol}
%\noindent \textbf{Detrac4reID Construction} 

\subsubsection{ Data Split. }
For model training and testing using the VRIC dataset as a benchmark, we randomly
split all 5,622 vehicle IDs into two non-overlapping halves: 
2,811 for training, and 2,811 for testing.
To remove data redundancy, 
we performed random frame-wise sub-sampling of the training trajectories.
%
%UA-DETRAC is a multi-tracking \emph{single}-camera detection and tracking 
%dataset unlike its counterparts in person tracking where identities observed
%across cameras.  
Since there is no cross-camera pairwise ID matches 
(UA-DETRAC is about single-camera object detection/tracking),
we simulated cross-view variation
by distant sampling between probe and gallery images.

In particular, we defined two pseudo views, {\em near} or {\em far},
for each video/camera and then built the probe/gallery 
sets from the test trajectories by randomly sampling each in two pseudo views.
It is shown in Fig \ref{fig:VRIC}(b) that
the {\em near} and {\em far} views present very different 
viewing conditions and hence allowing for 
a good simulation of two non-overlapping camera views.
In this sense, VRIC contains a total 120 pseudo camera views from the 
60 original camera views
with unconstrained condition diversity.

We adopted the standard single-shot evaluation setting,
i.e. one image per vehicle per view.
From the above, we obtained 54,808/5,622 training/testing images
for the VRIC benchmark.
The data partition and statistics are summarised
in Table \ref{table:data_split}.

\begin{table}[ht]
	\centering
%	\footnotesize
	\caption{Data statistics and partition in VRIC.}
%	\vspace{-0.2cm}
	\setlength{\tabcolsep}{0.5cm}
	\begin{tabular}{l||c||c|c|c}
		\hline 
		\multirow{2}{*}{Partition} 
		& \multirow{2}{*}{All}
		& \multirow{2}{*}{Training Set} & \multicolumn{2}{c}{Test Set} \\
		\cline{4-5}
		& & & Probe & Gallery  \\
		\hline \hline
		IDs
		& 5,622
		& 2,811 & 2,811 & 2,811 \\
		\hline
		Images 
		& 60,430 
		& 54,808 & 2,811 & 2,811 \\
		\hline
	\end{tabular}
	\label{table:data_split}
	%	\vspace{-.5cm}
\end{table}

%The resulting conversion has 56043 bounding boxed in
%training for 2828 identities and 5664 images for 2832 identities in testing.
%The bounding boxes are hxw pixels on average ranging from hxw to hxw with
%challenging multi scale variations. Test and
%training images are labelled with 120 camera views. 
%To alleviate 
%this shortcoming we have chosen to label a segmented moving vehicle's associated 
%camera by where it is in the original frame, i.e.~{\em near} or {\em far}. 
%By doing so we have labelled images with 120 cameras in Detrac4reID and each 
%vehicle is seen by two cameras. 

%The training set of UA-DETRAC yields roughly 
%300K images of 6K identities. Of
%these identities we eliminate identities with less than 20 frames 
%since we can't justify
%the viewpoint variety to separate those frames into 2 camera angles. Furthermore
%for each of the remaining identities $k$ number of frame indexes are randomly
%selected of an identity tracklet,  where $k$ is randomly selected integer between
%$1$ and $40$. This gives us a realistic setting with roughly 56043 images in the
%training set by randomly choosing half of the identities. 
%For the remaining half
%to ensure we have distinct camera angles for probe and gallery images, beginning
%and end frames of the tracklets in the test set are chosen as the probe and gallery
%image for that identity. Detrac4reID has 2784 and 2782 identities in train and 
%test set respectively. See Table~\ref{tbl:data-summary} for comparisons of all
%the datasets. AS we end up in a single-shot setting in Detrac4reID, evaluation
%is done with rank-1, rank-5 scores.

\subsubsection{  Performance Metrics. }
%Model performance is measured by {\em Cumulative Matching Characteristic} 
%(CMC) \cite{klare2015pushing} metric with Rank-k giving the proportion of test probes
%with the true match at rank $k$ or better, and 
For re-id performance measure, we used the {\em Cumulative Matching Characteristic}  (CMC) rates 
\cite{klare2015pushing}. The CMC is computed for each individual rank $k$ as
the cumulative percentage of the truth matches for probes returned
at ranks $\leq k$. 
In practice, the Rank-1 rate is often used as a strong indicator of an algorithm's efficacy.
%The mAP is to measure the recall of multiple
%truth matches, computed by first computing the area under
%the Precision-Recall curve for each probe, then calculating
%the mean of Average Precision over all probes.

\section{Deep Learning Multi-Scale Vehicle Representation}
\label{sec:method}
%\subsection{Problem Statement}
We aim to learn a deep representation model 
from a set of $n$ vehicle images 
$\mathcal{I} = \{ \bm{I}_i \}_{i=1}^n$ with the corresponding vehicle ID labels as $\mathcal{Y} = \{y_i\}_{i=1}^n$.
These training images capture the visual appearance
variations of 
$n_\text{id}$ different IDs
under multiple camera views,
with $y_i \in [1,\cdots, n_\text{id}]$.
In typical surveillance scenes,
vehicles are often captured at varying scales (resolutions),
which causes significant inter-view feature representation discrepancy in re-id matching.
In this work, we investigate this problem in vehicle re-id by
exploring image pyramid representation \cite{adelson1984pyramid,lazebnik2006beyond}.

Specifically, we exploit the potential of learning ID discriminative
pyramidal representations originally designed for person re-id
\cite{chen2018person}. 
Our objective is to extract and represent complementary appearance
information of vehicle ID from multiple resolution scales concurrently
in order to optimise re-id matching under significant view changes. 
We call this model {\bf Multi-Scale Vehicle Representation} (MSVR).
Our approach differs notably from
existing vehicle re-id models typically assuming single-scale
representation learning. 

\begin{figure}[ht]
	\centering
	\includegraphics[width=.97\linewidth]{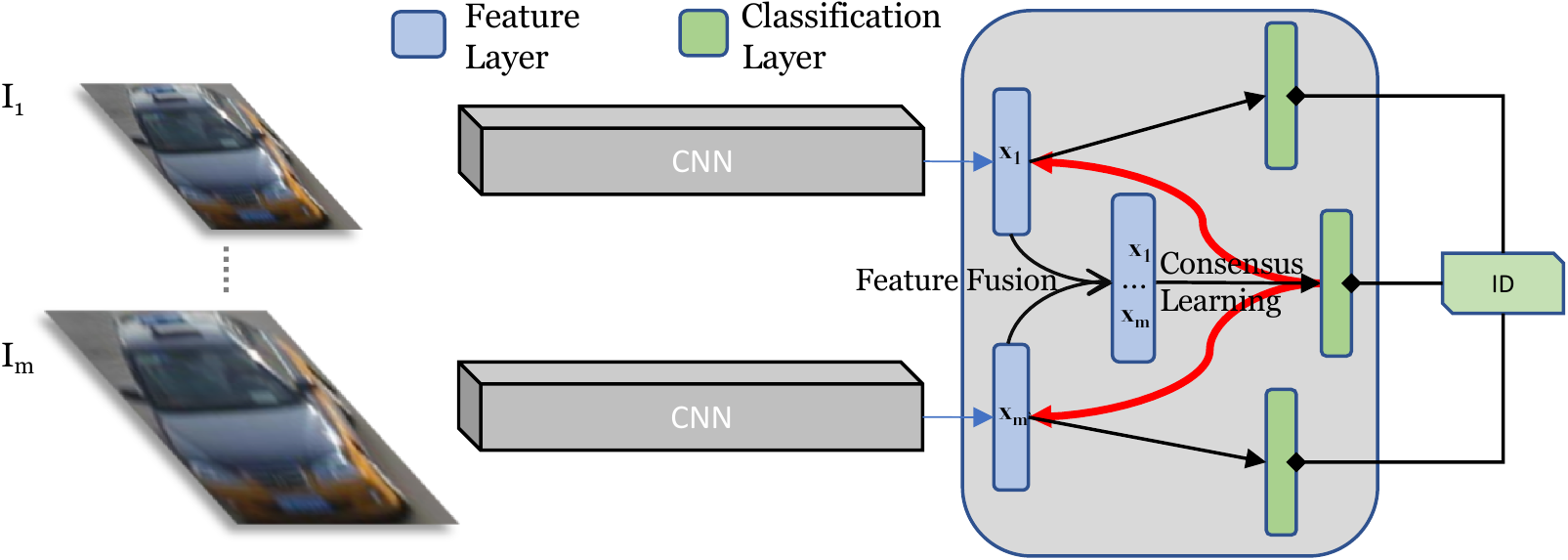}
%	\vspace{0.2cm}
	\caption{Overview of Multi-Scale Vehicle Representation (MSVR) learning
		for discriminative vehicle re-id at varying spatial resolutions.
		MSVR learns vehicle re-id sensitive feature representations
		from image pyramid by an network architecture of
        multiple branches all of which are optimised concurrently 
        (consensus feedback shown in {\color{red} red}, see Eq.~\eqref{equ:cross-entropy})
		subject to the same ID label constraints.
		Importantly, an inter-scale interaction mechanism
		is enforced to further enhance the scale-generic feature learning.
	% {\bf \color{red} TODO for Aytac. (Similar as Yanbei's but try to be different).}
}
	\label{fig:MSVR_pipeline}
%	\vspace{-0.2cm}
\end{figure}

\subsubsection{  MSVR Overview. }
The overall MSVR network design 
is depicted in Fig \ref{fig:MSVR_pipeline}. 
Specifically, MSVR consists of $(m+1)$ sub-networks:
(1) $m$ branches of sub-networks each for learning discriminative 
scale-specific visual features. 
Each branch has an identical structure.
(2) One fusion branch for learning the 
discriminative integration of $m$ scale-specific representations of 
the same vehicle image.
To maximise the complementary advantage between 
different scales of feature representation in learning,
we concurrently optimise
per-scale discriminative representations
with scale-specific and scale-generic (combined) learning
subject to the same ID label supervision. 
Critically, we further propagate multi-scale consensus 
as {feedback} to regulate the learning of per-scale branches. 
Next, we detail three MSVR components:
(1) Single-Scale Representation;
(2) Multi-Scale Consensus;
(3) Feature Regularisation.

% \vspace{0.2cm}
\subsubsection{  (1) Single-Scale Representation. }
We exploit the MobileNet \cite{howard2017mobilenets} 
to design single-scale branches 
due to its favourable trade-off
between model complexity and learning capability.
To train a single-scale branch,
we use the softmax cross-entropy loss function
to optimise vehicle re-id sensitive information from
ID labels.
Formally, we first compute the class posterior probability 
$\tilde{y}$ of a training image $\bm{I}$:

\begin{equation}
{p}(\tilde{y} = y | \bm{I}) = \frac{\exp(\bm{w}_{y}^{\top} \bm{x})} {\sum_{k=1}^{n_\text{id}} \exp(\bm{w}_{k}^{\top} \bm{x})}
\label{eq:prob}
\end{equation}
where $\bm{x}$ and $y$ refer to the feature vector and ground-truth label of $\bm{I}$,
$n_\text{id}$ the number of training IDs,
and $\bm{w}_k$ the classifier parameters
of class $k$.
The training loss is then defined as:

\begin{equation}
\mathcal{L}_\text{ce} = - \log \Big(p(\tilde{y} = y|\bm{I}) \Big)
\label{eq:loss_cls}
\end{equation}

% \vspace{0.2cm}
\subsubsection{  (2) Multi-Scale Consensus. }
We learn multi-scale consensus on vehicle ID classes between
$m$ scale-specific branches. 
We achieve this using joint-feature based classification.
First, we obtain joint feature of different scales by vector fusion.
In MobileNets, feature vectors are computed by 
global average pooling of the last CNN feature maps
with dimension of 1024.
Hence, this fusion produces a 1024$\times$$m$-D feature vectors.
We then use this combined features to 
perform classification for providing 
multi-scale consensus on the ID labels.
We again adopt the cross-entropy loss (Eq \eqref{eq:loss_cls})
as in single-scale representation learning.

% \vspace{0.2cm}
\subsubsection{  (3) Feature Regularisation. }
We regularise the single-scale branches
by multi-scale consensus for imposing interaction between
different scale representations in model learning.
Specifically, we propagate the consensus 
as an auxiliary {\em feedback} to regularise 
the learning of each single-scale branch concurrently.   
We first compute for each training sample
%Inspired by the {teacher-student} learning spirit~\cite{hinton2015distilling}, 
%we do this propagation by exploiting the sample-wise 
a soft probability prediction (i.e. a consensus representation)
$\tilde{P} = [\tilde{p}_1,\cdots,\tilde{p}_i,\cdots,\tilde{p}_{n_\text{id}}]$
as: 

\begin{eqnarray}
\tilde{p}_{i} = \tilde{p}(\tilde{y} = i | \bm{I}) = \frac{\exp(\frac{z_{i}}{T})}{\sum_{k} \exp(\frac{z_{k}}{T}) }
, \quad i \in [1,\cdots, n_\text{id}]
\label{equ:soft_prob}
\end{eqnarray}
where $z$ is the logit
and $T$ the temperature parameter (higher values leading to 
softer probability distribution).
We empirically set $T=1$ in our experiments.
Then, we use the consensus probability $\tilde{P}$ as the {\em teacher} signal 
to guide the learning process of each single-scale branch ({\em student}).
To quantify the alignment between these predictions,
we use the cross-entropy measurement which is defined as:

% between
%two distributions $\tilde{P}$ and $P$, i.e.
\begin{eqnarray}
\mathcal{H}(\tilde{P}, P) = - \frac{1}{n_\text{id}}
\sum_{i=1}^{n_\text{id}} \big( \tilde{p}_i \ln(p_i) + (1-\tilde{p}_i) \ln(1-p_i) \big)
\label{equ:cross-entropy}
\end{eqnarray}
The objective loss function for each single-scale branch is then:
 
\begin{eqnarray}
\mathcal{L}_\text{scale} = \mathcal{L}_\text{ce} + \lambda \mathcal{H}(\tilde{P}, P)
\label{equ:regul}
\end{eqnarray}
where the hyper-parameter $\lambda$ ($\lambda = 1$ in our experiments) is the weighting between two loss terms.
$P = [{p}_1,\cdots,{p}_{n_\text{id}}]$ defines the
probability prediction over all $n_\text{id}$ identity classes 
by
the corresponding single-scale branch (Eq. \eqref{eq:prob}).
As such, each single-scale branch learns to 
correctly predict the true ID label of training sample ($\mathcal{L}_\text{ce}$) 
by the corresponding scale-specific representation and 
to match the consensus probability estimated 
based on the scale-generic representation ($\mathcal{H}$).

% \vspace{0.2cm}
\noindent{\bf MSVR Deployment. }
In model test, 
we deploy the fusion branch's representation 
for multi-scale aware vehicle re-id matching.
We use only a generic distance metric without camera-pair specific
distance metric learning, e.g. the L2 distance.
Based on the pairwise distance, we 
then return a ranking of gallery images as the re-id results.
For successful tasks, the true matches for a given probe image
are should be placed among top ranks.

\section{Experiments}

%\subsection{Datasets and Settings}

\subsubsection{  Datasets. }
For evaluation, in addition to the newly introduced VRIC dataset,
we also utilised two most popular vehicle re-id benchmarks.
The {\bf VehicleID} \cite{Liu2016cvpr_drdl} dataset provides a training set with 113,346 from 13,164 IDs and a test set with 19,777 images from 2,400 identities.
It adopts the single-shot re-id setting, with only one true matching for each probe. Following the standard setting, 
we repeated 10 times of randomly selected probe and gallery sets
in our experiments. 
The {\bf VeRi-776} dataset~\cite{Liu2016eccv_veri} 
has 37,778 images of 576 IDs in training set
and 200 IDs in test set.
The standard probe and gallery sets consist of
1,678 and 11,579 images, respectively. 
The data split statistics are summarised in Table \ref{tbl:data-summary}.

\begin{table}[h]
	\centering
	\caption{Data split of vehicle re-id datasets evaluated in our experiments.}
	\label{tbl:data-summary}
%	\vspace{-0.2cm}
	%	\resizebox
	%	{\columnwidth}{!}
	{%
		\begin{tabular}{c||c|c|c}
			\hline
			Dataset                           
			& Training IDs / Images & Probe IDs / Images & Gallery IDs / Images   
			%			& Imgs per ID(Train) & Avg. Res.(Height/Width)  & Total Imgs.     
			\\
			\hline  \hline                                                                                                                     
			VehicleID\cite{Liu2016cvpr_drdl}    
			& 13,164 / 113,346        & 2,400 / 17,377       & 2,400 / 2,400    
			%			& 10.12             &  345.4/376.1    & 133123        
			\\ \hline
			VeRi-776\cite{Liu2016eccv_veri}      
			& 576 / 37,778           & 200 / 1,678         & 200 / 11,579  
			%			& 65.58             &  243.5/214.2    & 51035          
			\\ \hline
			VRIC ({\bf Ours})                       
			& 2,811 / 54,808         & 2,811 / 2,811        & 2,811 / 2,811       
			%			& 19.56             &  107.5/69.8     & 60032           
			\\    
			\hline
		\end{tabular}
	}
	\vspace{-0.3cm}
\end{table}

% \vspace{0.2cm}
\subsubsection{  Performance Metrics. } 
For VehicleID and VRIC, we used the CMC measurement to evaluate re-id performance.
For VeRi-776, we additionally 
adopted the {\em mean Average Precision} (mAP)
due to its multi-shot nature in the gallery of the test data.
Specifically, for each probe, we compute the area under its Precision-Recall curve,
i.e. Average Precision (AP). 
The mAP is then computed as the mean value of APs for all probes.
This metric considers both precision and recall performance, and hence providing a more comprehensive evaluation.

%Attribute labels for
%\textit{vehicle type} and \textit{colour} are also available. For evaluation
%rank-1, rank-5 and mAP scores are used.
%\noindent VehicleID~\cite{Liu2016cvpr_drdl}, VeRi-776~\cite{Liu2016eccv_veri} are 
%the datasets that have been used most in recent publications and we consider
%these datasets for evaluation. Both datasets are surveillance datasets that
%provide identity labels for vehicle re-id. VehicleID also provides attribute
%labels for \textit{ vehicle model } and \textit{ colour }. Training data is over
%113346 with 13164
%identities and the largest test set consist of 19777 images with 2400
%identities. For evaluation rank-1 and rank-5 metrics are calculated for 10 runs
%of randomly selected gallery size of 2400 (one image per identity) and 17377
%probe images. VeRi-776 has 37778 images of 576 identities in training set
%and in test set 200 identities with a predefined set of probe images of size
%1678 and 11579 images in gallery. VeRi-776 also provides complete camera labels
%for images as well as spatio temporal relations. Attribute labels for
%\textit{vehicle type} and \textit{colour} are also available. For evaluation
%rank-1, rank-5 and mAP scores are used.
%
%Total number of training images in both VeRi-776 and Detrac4reID similar, this
%gives a rigorous comparison with methods with CNNs where number of training data affects the
%results greatly and enables us to compare characteristics of the datasets
%thoroughly. 

% \vspace{0.2cm}
\subsubsection{  Implementation Details. }
In the MSVR model, we used 2 resolution scales,
$224\times224$ and $160\times160$.
We adopted the ADAM optimizer and 
set the initial learning rate to 0.0002, the
weight decay to 0.0002, the $\beta _1$ to 0.5,
the mini-batch size to 8,
the max-iteration to 100,000. 
Model initialization was done with ImageNet~\cite{imagenet} pretrained weights.
%,
%For our re-implementations of~\cite{Wang_2017_ICCV, Shen_2017_ICCV},
%We adopted the SGD with Nesterov momentum and 
%set the initial learning rate to 0.001, the
%weight decay to 0.0002, the momentum to 0.9,
%the mini-batch size to 128, 
%and the epoch to 60. %(SGD performed better than Adam optimizer). 
The data augmentation includes random cropping and horizontal flipping.

%All training is done for 60 epochs for all dataset
%with initial learning rate 0.001 for runs that initialize weights with ImageNet
%pre-trained weights and 0.01 otherwise. Stochastic Gradient Descent with
%\textit{ momentum } is the optimizer of choice in conjunction with lowering the
%learning rate every 40 epochs with batch size 128. Input images are resized to 224x244 pixels with random cropping and
%mirroring in the training phase. Center crops are used in the test phase. This is the general setting except otherwise
%noted. 

%\subsection{Comparative Evaluation}
% \vspace{0.2cm}
\subsubsection{  Evaluation. }
Table \ref{tbl:SOTA} compares MSVR with
state-of-the-art methods on three benchmarks.
We make these main observations as follows: \\
{\bf (1)}
Under the standard visual appearance based evaluation setting (the top part),
MSVR outperforms all other competitors with large margins --
MSVR surpasses the best competitor in Rank-1 rate
by 24.38 \% (88.56-64.18) on VeRi-776,
24.82\% (62.02-38.20) on VehicleID,
and 16.73\% (46.61-30.55) on VRIC.
This demonstrates the consistent superiority of MSVR 
over alternative methods in vehicle re-id, showing the importance in
modelling multi-scale representation for vehicle re-id. \\
{\bf (2)} 
Benefited from more training data plus space-time contextual knowledge and fine-grained
local key-point supervision, the OIFE model achieves the best performance
on VeRi-776. However, such advantages from additional data and
knowledge representation is generically beneficial to all models including the MSVR
when applied. \\
{\bf (3)} 
We carefully reproduced two very recent methods, OIFE(Single-Branch) \cite{Wang_2017_ICCV}  and Siamese-Visual \cite{Shen_2017_ICCV},
and obtained inconsistent results compared to the reported
performances of these two models.
In particular, the performance of OIFE(Single-Branch) decreases on VeRi-776 and VehicleID. This is mainly due to that the original results are based on
a larger multi-source training set with
225,268 training images of 36,108 IDs 
(from VehicleID~\cite{Liu2016cvpr_drdl}, 
VeRi-776~\cite{Liu2016eccv_veri}, BoxCars~\cite{sochor2016boxcars} and 
CompCars~\cite{yang2015compcars}), 
{\em versus} 
the standard 100,182 training images of 13,164 IDs on VehicleID, \ie
2.2 times more training images and 2.7 times more training ID labels,
and the standard 37,778 training images of 576 IDs on VeRi-776, \ie
6.0 times more training images and 62.7 times more training ID labels,
respectively.
In contrast, the result of Siamese-Visual (ResNet50 based) increases on VeRi-776.
It is worth pointing out that we trained this model using the cross-entropy classification loss
and cannot make it converge with pairwise inner-product loss.  

\begin{table}[h]
	\centering
	\setlength{\tabcolsep}{0.07cm}
	\caption{Comparative vehicle re-id results
		on three benchmarking datasets. Upper part of table lists methods trained with only the images available from the respective datasets for fair comparison of the methods; lower part lists methods trained with additional datasets and/or labels.  
		*: By our reimplementation. 
		%			P: ImageNet pretrained. 
		{\bf E}: Extra information and annotation, \eg~number plates, local key-points, space-time prior knowledge. 
		{\bf M}: Multiple vehicle re-id and classification datasets are combined for training.
        $\dagger$: Result from~\cite{Wang_2017_ICCV}.
        }
	\label{tbl:SOTA}
%	\vspace{-0.2cm}
	\resizebox{\columnwidth}{!}{%
		\begin{tabular}{c|c|cc|cc|cc|c}
			\hline
			\multirow{2}{*}{Method}
			& \multirow{2}{*}{\STAB{\rotatebox[origin=c]{90}{Notes}}} &  \multicolumn{2}{c|}{VeRi-776 \cite{Liu2016eccv_veri}} & \multicolumn{2}{c|}{VehicleID \cite{Liu2016cvpr_drdl}} & \multicolumn{2}{c|}{VRIC} & \multirow{2}{*}{Publication}        \\
			\cline{3-8}
			&                                                      
			& Rank-1   & mAP    
			& Rank-1   & Rank-5        
			& Rank-1   & Rank-5 &         \\ \hline \hline
%			 BOW-CN \cite{zheng2015scalable}                   &   
%			 & 33.91    & 12.20  
%			 & -        & -             
%			 & -        & -      & ICCV'15 \\
			 LOMO \cite{liao2015person}                        &   
			 & 25.33    & 9.64   
			 & -        & -             
			 & -        & -      & CVPR'15 \\
%			 GoogLeNet \cite{simonyan2014very}                 &   
%			 & 49.82    & 17.04  
%			 & -        & -             
%			 & -        & -      & CVPR'14 \\
			FACT \cite{Liu2016icme-veri}                        &   
			& 50.95     & 18.49  
			& -         & -             
			& -         & -      & ICME'16 \\
			Mixed Diff + CCL \cite{Liu2016cvpr_drdl}            &   
			& -         & -      
			& 38.20         & 50.30         
			& -      & -      & CVPR'16 \\
			Siamese-Visual \cite{Shen_2017_ICCV}                &   
			& 41.12    & 29.40  
			& -             & -             
			& -        & -      & ICCV'17 \\
%			Siamese-Visual (first stage) \cite{Shen_2017_ICCV}   & *
			Siamese-Visual \cite{Shen_2017_ICCV}   & *
			& 64.18    & 31.54  
			& 36.83    & 57.97         
			& 30.55    & 57.30 & ICCV'17 \\
			OIFE(Single Branch) \cite{Wang_2017_ICCV}           & * 
			& 60.13    & 31.81  
			& 32.86    & 52.75         
			& 24.62    & 50.98 & ICCV'17       \\
%			ResNet-50 \cite{he2016deep}         & P 
%			& 82.24    & 52.61  
%			& 45.26         & 73.13         
%			& 43.79  & 72.35    &         \\
%			\hline
			 \bf MSVF                                               &  
			 & \bf 88.56    &\bf 49.30
			 & \bf 63.02   & \bf 73.05             
			 & \bf 46.61  &\bf 65.58  & \bf Ours        \\
			 \hline
			 KEPLER \cite{martinel2015kernelized} $\dagger$        & M
			 & 68.70    & 33.53  
			 & 45.40    & 68.90             
			 & -        & -      & TIP'15  \\
			% FACT + Plate \cite{Liu2016eccv_veri}                &    & 61.08    & 77.41    & 25.88  & -             & -             & -      & -      & ECCV'16 \\
			FACT + Plate + Space-Time \cite{Liu2016eccv_veri}   & E  
			& 61.44    & 27.77  
			& -             & -             
			& -      & -     & ECCV'16 \\
			% Path-LSTM \cite{Shen_2017_ICCV}                     &    & 82.89    & 89.81    & 54.49  & -             & -             & -      & -      & ICCV'17 \\
			Siamese-CNN + Path-LSTM \cite{Shen_2017_ICCV}       & E  
			& 83.49    &\bf 58.27  
			& -        & -             
			& -        & -      & ICCV'17 \\
			OIFE(Single Branch) \cite{Wang_2017_ICCV}           & M  
			& 88.66    & 45.50  
			& 63.20    & 80.60         
			& -        & -     & ICCV'17 \\
			OIFE(4Views) \cite{Wang_2017_ICCV}                  & ME 
			& 89.43    & 48.00 
			&\bf 67.00    &\bf 82.90         
			& -        & -     & ICCV'17 \\
			OIFE(4Views + Space-Time) \cite{Wang_2017_ICCV}     & ME 
			&\bf 92.35    &51.42  
			& -        & -             
			& -        & -     & ICCV'17 \\
			\hline
		\end{tabular}
	}
%	\vspace{-0.3cm}
\end{table}

\subsubsection{  Further Analysis. }
Table \ref{tbl:scale_results} compares the 
performances of a single-scale and a multi-scale feature representations
of the MSVR model.
It is evident that the multi-scale representation learning with MSVR has
performance benefit across all three datasets with varying resolution scale changes.
This shows that the overall effectiveness of MSVR in boosting vehicle re-id matching performance.
Moreover, the model performance gain on VRIC is the largest,
which is consistent with the more significant
scale variations exhibited in the VRIC vehicle images (Fig \ref{fig:dataset_img} and Table \ref{tab:datasets}). 

\begin{table}[h]
	\centering
	\setlength{\tabcolsep}{0.25cm}
	\caption{Comparing single-scale and multi-scale representations of MSVR.
	Gain is measured as the performance difference of MSVR over the {\em mean} of single-scale variants.
	}
%	\vspace{-0.2cm}
	\label{tbl:scale_results}
	{%
		\begin{tabular}{c||cc|cc|cc}
			\hline
			Dataset
			& \multicolumn{2}{c|}{VeRi-776 \cite{Liu2016eccv_veri}} 
			& \multicolumn{2}{c|}{VehicleID \cite{Liu2016cvpr_drdl}} 
			% & \multicolumn{2}{c}{VRIC}   
			& \multicolumn{2}{c}{VRIC}   \\
			\hline
			Metrics (\%)                                                     
			& Rank-1   & mAP    
			& Rank-1   & Rank-5        
			% & Rank-1   & Rank-5        
			& Rank-1   & Rank-5        \\ \hline \hline
			Scale-224
			& 88.37 & 47.37
			& 62.80 & 72.54
			% & 45.13 & 62.82 
            & 43.55 & 61.88 \\
			Scale-160
			& 87.43 & 46.81
			& 60.29 & 71.15
			% & 45.27 & 62.82
            & 43.62 & 62.77
			\\ \hline
			MSVR                                                
			&\bf 88.56    &\bf 49.30
			&\bf 63.02  &\bf 73.05            
			% &\bf 48.02    &\bf 66.03 
            & \bf 46.61 & \bf 65.58 \\
			\hline \hline
			Gain (\%)
			& +0.76 & +2.11
			& +1.47 & +1.20
			& \bf +3.02 &\bf +3.25
			\\ \hline 
		\end{tabular}
	}
\end{table}

\section{Conclusion}

In this work we introduced a more realistic and challenging 
vehicle re-identification benchmark, Vehicle Re-Identification in
Context (VRIC), to enable the design and evaluation of vehicle re-id
methods to more closely reflect real-world application conditions.
VRIC is uniquely characterised by 
unconstrained vehicle images from large scale, wide scale traffic scene 
videos inherently exhibiting variations
in resolution, illumination, motion blur, and occlusion.
% and background clutter. 
This dataset provides a more realistic and truthful test and
evaluation of algorithms for vehicle re-id ``in-the-wild''. 
We further investigated a multi-scale learning representation
by exploiting a pyramid based deep learning method. 
Experimental evaluations demonstrate the effectiveness and performance
advantages of our multi-scale learning method over the
state-of-the-art vehicle re-id methods on three benchmarks VeRi-776, VehicleID,
and VRIC.
%a comprehensive and rigorous comparison of vehicle
%re-identification state of the  art methods using exactly the same data and
%preprocessing of images. Our experiments show that modern CNN architectures with
%standard techniques provide better scores without bells and whistles using only
%the visual information. We also introduced a new realistic multi-resolution
%vehicle re-identification dataset with characteristics more akin to its person
%re-identification counterparts highlighting more challenging setting backed by
%lower performance scores across all experiments. Finally an FPN inspired more
%resolution aware contruction of a model to improve our best results.

\section*{Acknowledgements}
This work was partially supported by
the Royal Society Newton Advanced Fellowship
Programme (NA150459), Innovate UK Industrial Challenge Project on Developing and Commercialising Intelligent Video Analytics Solutions for Public Safety (98111-571149),
Vision Semantics Ltd, and SeeQuestor Ltd.

% \clearpage

\bibliographystyle{splncs04}
\bibliography{bmvc}

\end{document}